\documentclass{article}

\usepackage[preprint, nonatbib]{nips_2018}

\usepackage[utf8]{inputenc} 
\usepackage[T1]{fontenc}    
\usepackage{hyperref}       
\usepackage{url}            
\usepackage{booktabs}       
\usepackage{amsfonts}       
\usepackage{nicefrac}       
\usepackage{microtype}      
\usepackage{graphicx}
\usepackage{subfiles} 
\usepackage{hyperref}
\usepackage[font=small,labelfont=bf]{caption}

\title{Towards Explainable Deep Learning for Credit Lending: A Case Study}

\author{
  Ceena Modarres \\
  Center for Machine Learning, Capital One \\
  \texttt{ceena.modarres@capitalone.com} \\
  \And
  Mark Ibrahim \\
  Center for Machine Learning, Capital One \\
  \texttt{mark.ibrahim@capitalone.com} \\
  \AND
  Melissa Louie \\
  Center for Machine Learning, Capital One \\
  \texttt{melissa.louie@capitalone.com} \\
  \And
  John Paisley \\
  \hspace{12mm} Columbia University \hspace{13mm} \\
  \texttt{jpaisley@columbia.edu} \\
}

\begin{document}

\maketitle

\begin{abstract}
Deep learning adoption in the financial services industry has been limited due to a lack of model interpretability. However, several techniques have been proposed to explain predictions made by a neural network. We provide an initial investigation into these techniques for assessment of credit risk with neural networks.\end{abstract}

\section{Deep Learning for Credit Risk Assessment}

Deep learning has transformed many areas of data science, but has experienced limited adoption in the financial industry, particularly for assessing credit worthiness. This not due to their performance, which has been shown to eclipse decision trees and logistic regression in credit risk related tasks, but rather a product of their black-box quality \cite{angelini2008neural} \cite{fei2015} \cite{stroie2013}. Both the Equal Credit Opportunity Act (ECOA), as implemented in Regulation B, and the Fair Credit Reporting ACT (FCRA) require lenders to provide reasons for denying a credit application \cite{adversea67}. Neural networks' black-box quality can hinder a lender's ability to assess model fairness, detect bias, and meet regulatory demands.

Unlike linear models, where the learned feature weights indicate relative importance, there is no clear mechanism for assessing the reasons behind a neural network's prediction \cite{doshi2017towards}. Several attribution techniques have been proposed to explain neural network predictions, but have not significantly filtered into the financial industry \cite{NIPS2017_7062} \cite{frosst2017distilling}. An attribution technique explains a given prediction by ranking the most important features used for generating that prediction. Three prominent techniques are LIME \cite{ribeiro2016should}, DeepLift \cite{shrikumar2017learning}, and Integrated Gradients \cite{sundararajan2017axiomatic}. These methods are theoretically founded and provide human interpretable reasons for a prediction. However, they have not yet been studied or applied in the context of credit assessment. In order to evaluate the viability of attribution methods in credit lending, we ground our analysis in evaluating two important questions:

\begin{enumerate}
\itemsep0em 
  \item Do these attribution methods generate accurate and interpretable explanations for a credit decision? (Trustworthiness)
  \item How consistently do attribution methods produce \textit{trustworthy} explanations for a credit decision? (Reliability)
\end{enumerate}

We explore the trustworthiness and reliability of LIME, DeepLift, and Integrated Gradients and propose two future research directions for the practical use of attribution methods in credit lending \cite{kindermans2017reliability} \cite{adebayo2018local} \cite{adebayo2018sanity}.

\section{Background: Explaining Credit Decisions}

We consider FICO's open-source Explainable Machine Learning dataset of Home Equity Line Credit applications.
\cite{explaina31}.
The dataset provides information on credit candidate applications and risk outcome ($Y$) - characterized by 90-day payment delinquencies ($Y \in \{0,1\}$). 
To test the reliability and trustworthiness of attributions, we train a feed forward neural network to forecast credit risk
(dataset and model details in \textbf{Appendix}).

\section{Experiment 1: Trustworthiness of Attributions for Credit Lending}

\paragraph{Generating an Accepted Ground Truth}
We train a logistic regression classifier to predict credit risk.
We then measure feature importance through the learned feature weights. 
We treat the learned feature importance as the accepted ground truth explanation due to its wide application in credit risk modeling by practitioners, acceptance among regulatory bodies, and theoretical grounding 
\cite{Creditsc67}.
We train a logistic regression on the FICO dataset and use the learned global weights as the benchmark to compare explanations of neural network predictions.

\paragraph{Neural Network Explanations}
We generate explanations for the neural network model using LIME, DeepLIFT, and Integrated Gradients. 
LIME produces local explanations by perturbing the input around a neighborhood and fitting a linear model. 
Integrated Gradients and DeepLIFT are gradient-based methods that use a baseline vector to identify the feature dimensions with the highest activations \cite{ancona2018towards}.

\begin{figure}[h]
\centering
\includegraphics[width=1.0\textwidth]{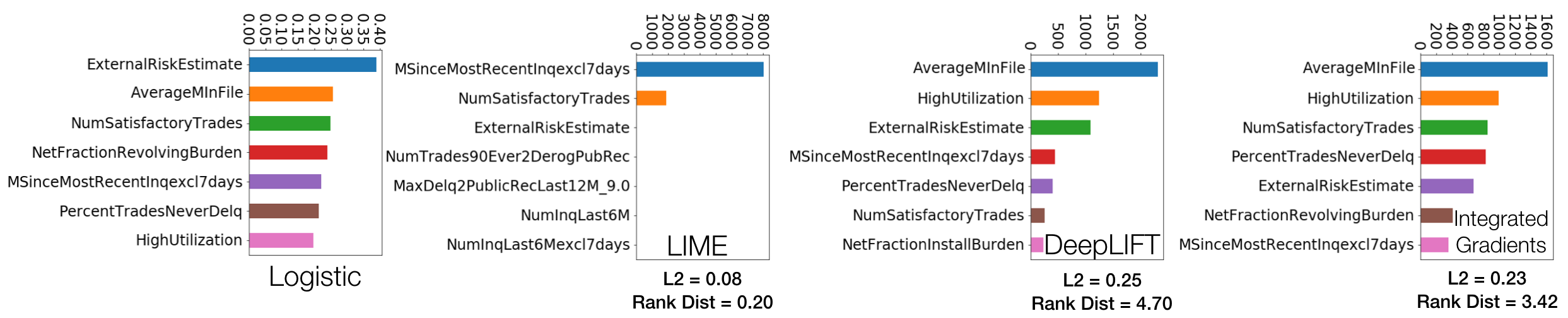}
\caption{Top 7 features for each technique evaluated by absolute weight (Logistic) or proportion of samples ranked by importance (Attribution). L2 and Rank Dist measure the average L2 norm and Spearman's rho weighted rank distance between local explanations and global weights.}

 \label{fig:featureImportance}
\end{figure}

We apply all three attributions techniques on the trained neural network using a neutral baseline ( $\sigma(r) \approx [0.50, 0.50]$) (see ~\ref{appendix} for details). 
We compare the attributions against the accepted global feature importance in two ways: 1) concordance of the top features 2) similarity to the accepted global weights (see Figure ~\ref{fig:featureImportance}) \cite{Lee12}.

Experimental results are displayed in Figure ~\ref{fig:featureImportance}.
Both DeepLIFT and Integrated Gradients produce similar top features. 
Of logistic regression's top 7 features, DeepLIFT agrees on 6 and Integrated Gradients agrees on all 7 features. However, the ranking varies across all three methods.
On average, Integrated Gradients produces explanations closer to the global weights in terms of rank and L2 (see Figure ~\ref{fig:featureImportance}).
Although LIME has a smaller distance to the accepted global explanation, it only identifies two of the top 7 features. This skewed distribution accounts for the smaller norms, but indicates that LIME fails to capture the full set of predictive features.

\subsection{Future Research Direction - Attribution Discrepancy}

Despite notable agreement, there is discrepancy across all three techniques between generated and accepted explanations. 
In order for practitioners to trust explanations of neural network credit risk predictions, we propose a future research direction: 
\textit{Is the discrepancy in model explanations due to differences in the learned interactions or due to error in the attributions?}

One approach would dissect the decision boundary learned by neural networks to compare against the feature interactions learned by a surrogate model.
In addition, the research could explore a measure of an attribution's fidelity to an accepted explanation. This would require a systematic approach by which to compare local attributions against an accepted global explanation.

\section{Experiment 2: Reliability of Attribution Methods}

The Integrated Gradients and DeepLIFT algorithms
require the specification of a reference point or baseline vector ($r$). The explanation of a credit decision is created relative to
the selected reference point.
Some domains have a clear cut reference - in image processing it is standard to use an all black image.
In credit lending, there is no context-specific baseline. 

In this section,
we explore the impact of the reference point on explanations in credit risk modeling (reliability), evaluate possible credit-lending specific reference points, 
and propose future research directions. 

\subsection{Experiment - Reliability of Attribution Methods to Baseline Selection}

We explore the reliability
of both Integrated Gradients and DeepLift to selecting a random baseline. This reference point should serve as a basis to explain 
a credit risk prediction. 
We use this as a foundation to evaluate the reliability of each explanation method relative to the choice of reference point.

To evaluate reliability, we generate three sets of references ($\{r_j\}_{j=1}^{K}$). In the first set, we randomly generated baselines between the minimum and maximum value
of each feature in our FICO dataset. In the second set, we interpret the recommendation of a neutral input in Sundararajan et al., 2017 to mean candidates 
that lie on the decision boundary ($\sigma(r_j) \approx [0.50, 0.50]$) \cite{sundararajan2017axiomatic}. For the third set, we tightly constrain generated reference points to
10 closest candidates on the decision boundary.

\begin{table}[h]
\centering
\begin{tabular}{|c|cc|cc|cc|}
\hline
Baseline 
& 
\multicolumn{2}{c|}{Random}    
&                                           
\multicolumn{2}{c|}{Constrained}
&
\multicolumn{2}{c|}{Tightly Constrained} \\

Attribution Method & Entropy & Std Deviation  & Entropy  & Std Deviation  & Entropy  & Std Deviation \\
\hline
IG & 0.121 & 0.002 & 0.115 & 0.001 & 0.192 & 0.001 \\
DeepLift & 0.119 & 0.002 & 0.103 & 0.001 & 0.230 & 0.000 \\
\hline
\end{tabular}
  \caption{Attribution baseline sensitivity experimental results. Attribution uncertainty metrics evaluated across
  		three runs. One with totally random baselines, another with heuristically constrained
		baseline and a third that is tightly constrained. Results imply that a tight constraint reduces uncertainty.}
  \label{entrop-exper}
\end{table}

\begin{figure}[h]
\centering
\includegraphics[scale=.55]{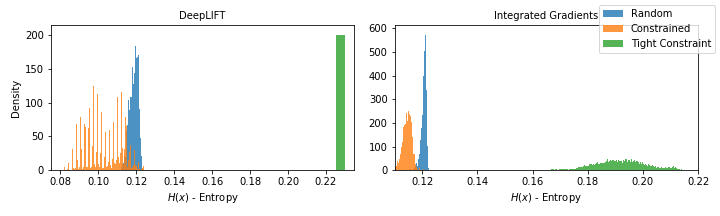}
\caption{Histogram of entropy values for credit candidates with varying reference points ($r_j$). Tightly constrained reference points reduce uncertainty of attributions 
across reference values. This result suggests that attribution methods are sensitive to the reference point and there may be desirable properties associated with
 varying the reference for each credit candidate.}
 \label{entropyHist}
\end{figure}

For a single candidate $(x)$, we generate $K$ attributions --- varying the reference
for each attribution ($A_{r_j}(x)$). We then evaluate the uncertainty of attributions for each candidate with two metrics: standard
 deviation and Shannon entropy defined as
 
\begin{equation}
  \label{entropy}
H(x) = -\frac{1}{K}\sum_{j=1}^{K} A_{r_j}(x)\ln (A_{r_j}(x)).
\end{equation}

We treat a feature's attribution across a varied reference as a probability density, so more uniform attributions correspond to higher entropies 
(see \textbf{Appendix}).
 Experimental findings are displayed in Table ~\ref{entrop-exper} and Figure ~\ref{entropyHist}.

We found that varying the reference point can have as large of an impact on a candidate's explanation as the applicant's actual profile. 
However, when references are constrained to only input samples that lie on the decision boundary, 
there is a significant decrease in the standard deviation of an explanation
 for a single candidate across both attribution methods. When we constrain references to only the ten most similar candidates on the decision boundary,
  there is an even larger reduction in dispersion and entropy increases. This result suggests that there may be some variation in the most appropriate reference 
  point for different candidates.
 
\subsection{Intuitive Baselines in Credit Lending are not on the Decision Boundary}

To shed light on reference selection for attribution methods in credit lending, 
we consider three potential candidate references: a candidate that is on the decision boundary, a candidate with no credit history, 
and the average candidate in our dataset. The results are shown in Table ~\ref{referenceProfile}.

\begin{table}[h]
\begin{center}
\begin{tabular}{ |c|c|c|c| } 
 \hline
   & Unclassifiable & Avg Candidate & New Candidate  \\
    \hline
  $\sigma(r)$ & $[0.50, 0.50]$ & [0.41, 0.59] & $[0.00, 1.00]$ \\
    \hline
     Delinquencies & 1  & 1 & 0\\
    Credit History (years) & 16 & 16 & 0 \\
  Credit Trades* & 21 & 20 & 0 \\
  Accts with Balance & 78\% &  67\% & NA\\
  Credit Inquiries (Last 6mo) & 2 & 1 & 0\\
\hline
\multicolumn{4}{|c|}{*Credit trades refers to any agreement between a lending agency and consumers} \\
\hline
\end{tabular}
  \caption{Credit profiles of potential reference points.}
  \label{referenceProfile}
  \end{center}
\end{table}

We see a notable discrepancy between qualitatively intuitive baselines and the softmax output. These candidates would ideally 
serve as a neutral point to frame insightful and actionable explanations for a credit decision. However, our model evaluates these candidates as high risk. A candidate that lies on the 
decision boundary, on the other hand, has no significance in credit lending. Ultimately, this does not mean intuitive options 
cannot or should not be used. In the following section we propose a research direction for understanding reference selection in credit lending.

\subsection{Future Research Direction - Reference Point User Study}
 
We have shown that attribution methods are sensitive to the choice of reference and that there is no obvious reference in credit lending. On the one hand, intuitive references
provide a foundation upon which we can produce explanations (e.g. the candidate has more credit trades than the average customer). On the other hand, these intuitive
references may not be capable of providing actionable suggestions.
Thus, it is important to understand \textit{which reference is most understandable, trustworthy, and reliable for the user}. 

We propose a research study that explores 
the reaction and comprehension of a credit candidate to many different types of explanations \cite{DBLP2018}. Specifically, we want to compare suggestion-based explanations
with explanations relative to intuitive references. We found empirical evidence that personalizing references increases the reliability of explanations. Therefore, we want to additionally consider personalization by providing candidate-specific references that describe an individual's best path to 
credit worthiness. 
\section{Conclusion}

For the successful adoption of advanced machine learning algorithms in credit markets, model interpretability and explainability is necessary to meet regulatory demands and ensure fair lending practices. In this paper, we explored the process of explaining credit lending decisions made by a neural network using three different attribution methods: LIME, DeepLIFT, and Integrated Gradients. Using the widely-accepted logistic regression as a surrogate for ground truth interpretations, we found some evidence that attribution methods are reliable. We also found that attribution methods are sensitive to the selected reference point. We discussed future research direction to help make deep learning perceived as more trustworthy and reliable by those working in credit lending.

\pagebreak

\small{\textbf{Disclaimer:} the opinions in the paper are the opinions of the authors and not Capital One.}

\bibliographystyle{plain}

\begin{thebibliography}{10}

\bibitem{adebayo2018local}
Julius Adebayo, Justin Gilmer, Ian Goodfellow, and Been Kim.
\newblock Local explanation methods for deep neural networks lack sensitivity
  to parameter values.
\newblock {\em ICLR 2018 Workshop}, 2018.

\bibitem{adebayo2018sanity}
Julius Adebayo, Justin Gilmer, Michael Muelly, Ian Goodfellow, Moritz Hardt,
  and Been Kim.
\newblock Sanity checks for saliency maps.
\newblock {\em arXiv preprint}, 2018.

\bibitem{adversea67}
Sarah Ammermann.
\newblock Adverse action notice requirements under the ecoa and the fcra -
  consumer compliance outlook.
\newblock
  \url{https://consumercomplianceoutlook.org/2013/second-quarter/adverse-action-notice-requirements-under-ecoa-fcra/},
  2013.
\newblock Federal Reserve Bank of Minneapolis.

\bibitem{ancona2018towards}
Marco Ancona, Enea Ceolini, Cengiz Öztireli, and Markus Gross.
\newblock Towards better understanding of gradient-based attribution methods
  for deep neural networks.
\newblock In {\em International Conference on Learning Representations}, 2018.

\bibitem{angelini2008neural}
Eliana Angelini, Giacomo di~Tollo, and Andrea Roli.
\newblock A neural network approach for credit risk evaluation.
\newblock {\em The quarterly review of economics and finance}, 48(4):733--755,
  2008.

\bibitem{doshi2017towards}
Finale Doshi-Velez and Been Kim.
\newblock Towards a rigorous science of interpretable machine learning.
\newblock {\em arXiv preprint arXiv:1702.08608}, 2017.

\bibitem{explaina31}
FICO.
\newblock Explainable machine learning challenge.
\newblock
  \url{https://community.fico.com/s/explainable-machine-learning-challenge},
  2018.

\bibitem{frosst2017distilling}
Nicholas Frosst and Geoffrey Hinton.
\newblock Distilling a neural network into a soft decision tree.
\newblock {\em arXiv preprint arXiv:1711.09784}, 2017.

\bibitem{fei2015}
Samsul Islam, Lin Zhou, and Fei Li.
\newblock Application of artificial intelligence (artificial neural network) to
  assess credit risk: A predictive model for credit card scoring, 04 2015.

\bibitem{kindermans2017reliability}
Pieter-Jan Kindermans, Sara Hooker, Julius Adebayo, Maximilian Alber, Kristof~T
  Sch{\"u}tt, Sven D{\"a}hne, Dumitru Erhan, and Been Kim.
\newblock The (un) reliability of saliency methods.
\newblock {\em arXiv preprint arXiv:1711.00867}, 2017.

\bibitem{Lee12}
Paul~H Lee and LH~Philip.
\newblock Mixtures of weighted distance-based models for ranking data with
  applications in political studies.
\newblock {\em Computational Statistics \& Data Analysis}, 56(8):2486--2500,
  2012.

\bibitem{NIPS2017_7062}
Scott~M Lundberg and Su-In Lee.
\newblock A unified approach to interpreting model predictions.
\newblock In I.~Guyon, U.~V. Luxburg, S.~Bengio, H.~Wallach, R.~Fergus,
  S.~Vishwanathan, and R.~Garnett, editors, {\em Advances in Neural Information
  Processing Systems 30}, pages 4765--4774. Curran Associates, Inc., 2017.

\bibitem{DBLP2018}
Menaka Narayanan, Emily Chen, Jeffrey He, Been Kim, Sam Gershman, and Finale
  Doshi{-}Velez.
\newblock How do humans understand explanations from machine learning systems?
  an evaluation of the human-interpretability of explanation.
\newblock {\em CoRR}, abs/1802.00682, 2018.

\bibitem{ribeiro2016should}
Marco~Tulio Ribeiro, Sameer Singh, and Carlos Guestrin.
\newblock Why should i trust you?: Explaining the predictions of any
  classifier.
\newblock In {\em Proceedings of the 22nd ACM SIGKDD international conference
  on knowledge discovery and data mining}, pages 1135--1144. ACM, 2016.

\bibitem{ross2014mutual}
Brian~C. Ross.
\newblock Mutual information between discrete and continuous data sets.
\newblock {\em PLOS ONE}, 9(2):1--5, 02 2014.

\bibitem{shrikumar2017learning}
Avanti Shrikumar, Peyton Greenside, and Anshul Kundaje.
\newblock Learning important features through propagating activation
  differences.
\newblock {\em arXiv preprint arXiv:1704.02685}, 2017.

\bibitem{Creditsc67}
Nikos Skantzos.
\newblock Credit scoring: Case study in data analytics.
\newblock
  \url{https://www2.deloitte.com/content/dam/Deloitte/global/Documents/Financial-Services/gx-be-aers-fsi-credit-scoring.pdf},
  2016.
\newblock Deloitte Financial Services.

\bibitem{stroie2013}
Laura~Badea Stroie.
\newblock Techniques for customer behaviour prediction: A case study for credit
  risk assessment.
\newblock
  \url{https://ec.europa.eu/eurostat/cros/system/files/NTTS2013fullPaper_171.pdf}.

\bibitem{sundararajan2017axiomatic}
Mukund Sundararajan, Ankur Taly, and Qiqi Yan.
\newblock Axiomatic attribution for deep networks.
\newblock {\em arXiv preprint arXiv:1703.01365}, 2017.

\end{thebibliography}

\section{Appendix}
\label{appendix}

\subsection{Mapping to Original FICO Feature Names}
\begin{tabular}{ |l|l| } 
    \hline
    Raw Feature Name & Description \\
    \hline
RiskPerformance	& Paid as negotiated flag (12-36 Months) \\
ExternalRiskEstimate & Consolidated version of risk markers\\
MSinceOldestTradeOpen & Since Oldest Trade Open\\
MSinceMostRecentTradeOpen & Since Most Recent Trade Open\\
AverageMInFile & Average Months in File\\
NumSatisfactoryTrades &	Number Satisfactory Trades\\
NumTrades60Ever2DerogPubRec & Number Trades 60+ Ever\\
NumTrades90Ever2DerogPubRec & Number Trades 90+ Ever\\
PercentTradesNeverDelq & Percent Trades Never Delinquent\\
MSinceMostRecentDelq & Months Since Most Recent Delinquency\\
MaxDelq2PublicRecLast12M & Max Delq/Public Records Last 12 Months\\
MaxDelqEver	& Max Delinquency Ever. See tab "MaxDelq" for each category\\
NumTotalTrades & Number of Total Trades (total number of credit accounts)\\
NumTradesOpeninLast12M & Number of Trades Open in Last 12 Months\\
PercentInstallTrades & Percent Installment Trades\\
MSinceMostRecentInqexcl7days & Months Since Most Recent Inq excl 7days\\
NumInqLast6M & Number of Inq Last 6 Months\\
NumInqLast6Mexcl7days & Number of Inq Last 6 Months excl 7days\\
NetFractionRevolvingBurden & Net Fraction Revolving Burden\\
NetFractionInstallBurden & Net Fraction Installment Burden\\
NumRevolvingTradesWBalance & Number Revolving Trades with Balance\\
NumInstallTradesWBalance & Number Installment Trades with Balance\\
NumBank2NatlTradesWHighUtilization & Number Bank/Natl Trades w high utilization ratio\\
PercentTradesWBalance & Percent Trades with Balance\\
\\
\hline
\end{tabular}

\subsection{Mutual Information}

We also measure the joint mutual information of each feature with the target variable. Mutual information quantifies the informational content obtained from one variable through another. It is commonly 
used for feature selection to determine which set of feature is important for a particular task. 
We find the top features via global weights learned by logistic regression correspond to the features with higher mutual information.

In the context of a binary target $Y$ and a feature $X_i$, we measure joint mutual information as

$$ 
I(X_i, Y) = 
\int_{X_i}  \sum_Y p_{x_i, y}(x_i, y) \log \frac{p_{x_i, y}(x, y)}{p_{x_i}(x_i)p_y(y)}\ dx 
$$

where $p_{x_i}(x_i), p_y(y)$ are marginal densities. If $X_i$ and $Y$ are unrelated, then by independence $p_{x,y}(x, y) = p_x(x)p_y(y)$, implying 
0 mutual information. The stronger the association between $X_i$ and $Y$, the larger the mutual information.
Mutual information can be interpreted as the entropy in feature $X_i$ minus the entropy in $X_i | Y$. 

Our goal is to identify the features with the maximal joint mutual information to our target. This should correspond to the most important variable. We do so by selecting most important feature by maximizing joint mutual information: $ I(X_i, Y) $ for each feature $X_i$.  Here we estimate mutual information to following the approach described Ross et al. 2014 \cite{ross2014mutual}. We find the features with the high mutual information matches the top feature importances ascertained via logistic regression weights.

\subsection{Logistic Regression Classifier Training}

We train a logistic regression binary classifer to predict credit risk on the FICO dataset. We ran grid search on the regularization parameter using both l1 and l2 norms. We withhold 33\% of samples for validation and achieve a validationa accuracy of 73\% on the hold-out set. The dataset has balanced classes.

\subsection{Neural Network Model Architecture}

We train a neural network binary classifier to predict credit risk on the FICO dataset. The model uses 

\begin{itemize}
  \item 2 hidden layers with 17 and 5 RELU activation units
  \item sigmoid activation for output probability
  \item trained for 20 epochs with batch size of 100 samples
\end{itemize}

The model achieves 73\% validation accuracy on the balanced FICO dataset with (1/3 of samples held out for validation). 

\subsection{Attribution Techniques Background}
\textbf{LIME} produces interpretable representations of complicated models by optimizing two metrics: interpretability $\Omega$ and local fidelity $\mathcal{L}$. Defined as,

\begin{equation}
\zeta(x) ={\mathrm{argmin}}_{g \in G} \mathcal{L}(f, g, \pi_x) + \Omega(g).
\label{eq:lime}
\end{equation}

The comprehensibility $\Omega$ of an interpretable function $g\in G$ (e.g. non-zero coefficients of a linear model) is optimized alongside the fidelity $\mathcal{L}$ or faithfulness of $g$ in approximating true function $f$ in local neighborhood $\pi_x$.

In practice, an input point $x$ is perturbed by random sampling in a local neighborhood and a simpler linear model is fit with the newly constructed synthetic data set. The method is model agnostic, which means it can be applied to neural networks or any other uninterpretable model. The now explainable linear model's weights can be used to interpret a particular model prediction. Here we use the implementation of LIME provided by the open-source LIME python package.

\textbf{Integrated Gradients} is founded on axioms for attributions. First, sensitivity states that for any two inputs that differ in only a single feature and have different predictions, the attribution should be non-zero. Second, implementation invariance states that two networks that have equal outputs for all possible inputs should also have the same attributions. In this paper, the authors develop an approach that takes the path integral of the gradient for a particular point $x_i$ and the model's inference $F$ on the path ($\alpha$) between a zero information baseline $x'$ and the input $x$. 

$$IntegratedGrads_i(x) ::= (x_i - x'_i) \times \int_{\alpha=0}^1 \frac{\partial f(x'+\alpha \times (x-x'))}{\partial x_i}  d\alpha$$

The path integral between the baseline and the true input can be approximated with Riemman sums. An attribution vector for a particular observation (local) is produced as a result. The authors found that 20 to 300 steps can sufficiently approximate the integral within 5\%.

\textbf{DeepLIFT} explains the difference in output $\Delta t$ between a 'reference'  or 'baseline' input $t^0$ and the original input $t$ for a neuron output: $\Delta t = t - t^0$. For each input $x_i$, an attribution score is calculated $C_{\Delta x_i \Delta t}$ that should sum up to the total change in the output $\Delta t$. DeepLIFT is best defined by this described summation-to-delta property: $$\sum_{i=1}^{n} C_{\Delta x_i \Delta t} = \Delta.t$$ The author use a function analogous to a partial derivative and use the chain rule to trace the attribution score from the output layer to the original input. This method also requires a reference vector.

\subsection{On Entropy Score in Experiment 2}
In experiment 2, we state that a lower uncertainty corresponds to a higher entropy. For a probability density function, a higher uncertainty is typically associated with a higher entropy. However, we normalize attributions and use them as a probability density function instead of fitting a distribution to the data. Because we made this choice, a set of very similar attributions across a varied reference would have high entropy. Consider the toy example of a feature attribution across three different references which produced uniform values of $[1,1,1]$. We would normalize this set of feature attributions to $[.33,.33,.33]$ and treat this as the probability density. Since this density represents a uniform distribution, it would produce maximal entropy. 

\subsection{Feature Importance and Attributions}

\begin{figure}[h]
 \label{Feature-Importance}
\centering
\includegraphics[scale=.5,width=1.0\textwidth]{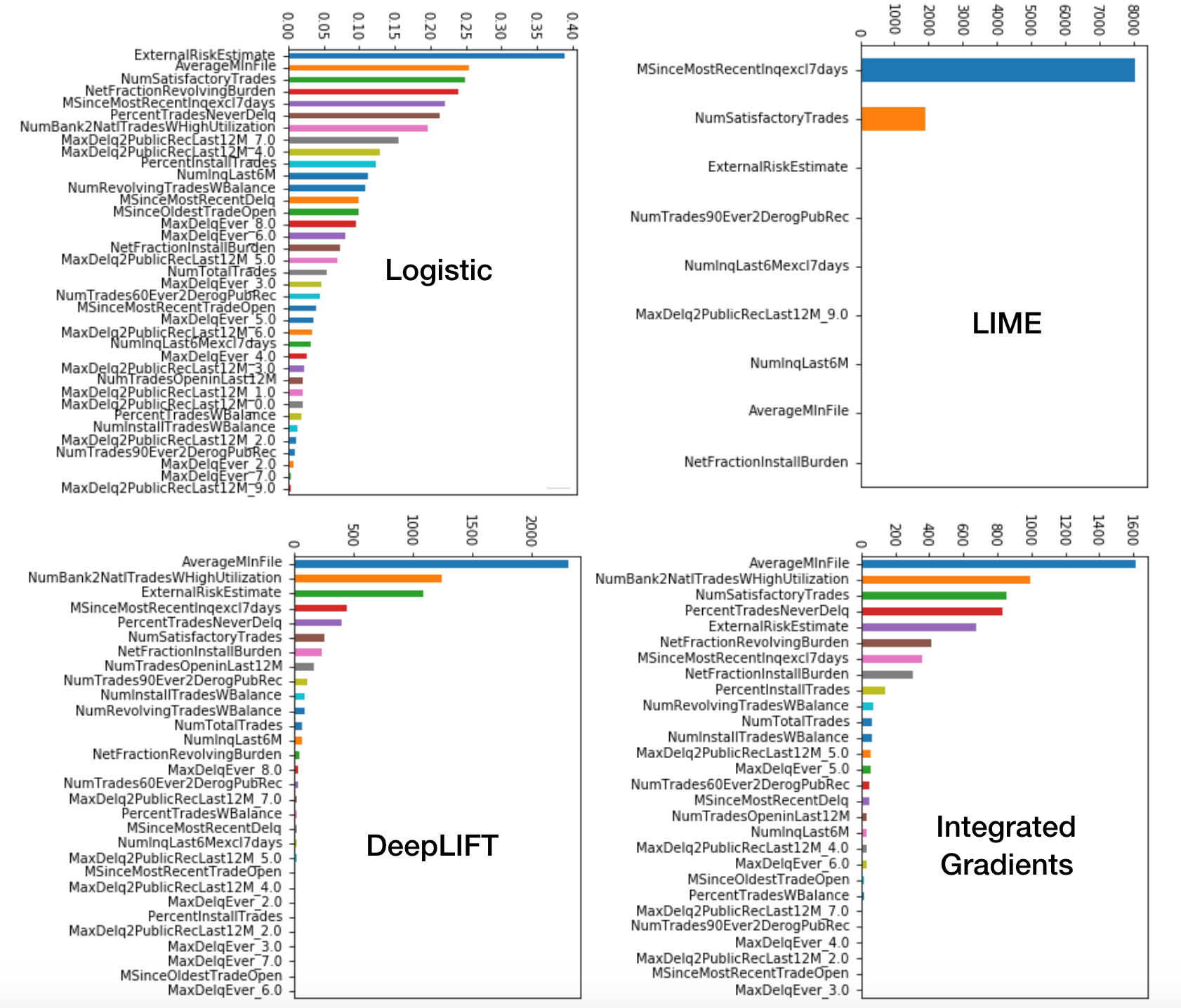}
\caption{Feature Importance for each technique. Learned absolute feature weights are used for logistic regression and frequency ranking across all samples is used for LIME, DeepLIFT, and Integrated Gradients.}
\end{figure}

\end{document}